\documentclass[10pt,twocolumn,letterpaper]{article}

\usepackage{wacv}
\usepackage{times}
\usepackage{epsfig}
\usepackage{graphicx}
\usepackage{amsmath}
\usepackage{amssymb}
\usepackage{booktabs}
\usepackage{subcaption}
\usepackage{lipsum}

\usepackage[accsupp]{axessibility}  

\usepackage{multirow}
\usepackage{bbold}
\usepackage{tabularx}
    \newcolumntype{L}{>{\centering\arraybackslash}X}

\usepackage{pifont}
\newcommand{\xmark}{\ding{55}}%

\def\wacvPaperID{198} 

\wacvalgorithmstrack   

\wacvfinalcopy 

\ifwacvfinal
\usepackage[breaklinks=true,bookmarks=false]{hyperref}
\else
\usepackage[pagebackref=true,breaklinks=true,colorlinks,bookmarks=false]{hyperref}
\fi

\usepackage{cleveref}
\begin{document}

\title{CTrGAN: Cycle Transformers GAN for Gait Transfer}
\author{Shahar Mahpod ~~~~~ Noam Gaash ~~~~~ Hay Hoffman ~~~~~  Gil Ben-Artzi \\
Ariel University \thanks{This research has been supported by the Ariel Cyber Innovation \mbox{Center}. Computing resources for this research were provided by Ariel University's HPC Center.} \\
 Israel \\
 {\tt\small \url{http://gil-ba.com} }
}


\maketitle
\thispagestyle{empty}

\begin{abstract}
We introduce a novel approach for gait transfer from unconstrained videos in-the-wild. In contrast to motion transfer, the objective here is not to imitate the source's motions by the target, but rather to replace the walking source with the target, while transferring the target's typical gait. Our approach can be trained only once with multiple sources and is able to transfer  the gait of the target from unseen sources, eliminating the need for retraining for each new source independently. Furthermore, we propose novel metrics for gait transfer based on gait recognition models that enable to quantify the quality of the transferred gait, and show that existing techniques yield a discrepancy that can be easily detected. 

We introduce Cycle Transformers GAN (CTrGAN), that consist of a decoder and encoder, both Transformers, where the attention is on the temporal domain between complete images rather than the spatial domain between patches. Using a widely-used gait recognition dataset, we demonstrate that our approach is capable of producing over an order of magnitude more realistic personalized gaits than existing methods, even when used with sources that were not available during training. As part of our solution, we present a detector that determines whether a video is real or generated by our model.
\end{abstract}

\section{Introduction}

The goal of this paper is to provide both a novel approach and a detection mechanism for gait transfer from videos in-the-wild. The objective is to replace a walking person (source) in a video sequence with photorealistic images of a different walking person (target), such that the resulting gait is identifiable as the target's while still mimicking the source's basic motion. 

Our approach learns directly from an unpaired collection of unconstrained videos in-the-wild containing walking people. We avoid the need for paired data and the need to disentangle the walking patterns into different representations and learn directly from the 2D frames. We train our model to translate multiple sources to a single target, so at inference time it can generalize to unseen sources without the need for retraining. 
 


\begin{figure*}	
	\begin{center}
	\includegraphics[width=1\textwidth]{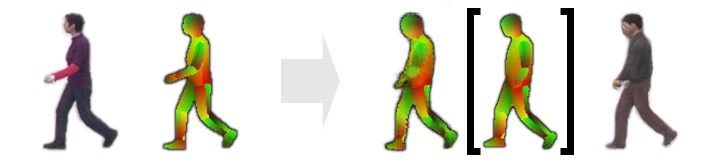}\\
	(a)~~~~~~~~~~~~~~~~~~~~~~~~~~~~(b)~~~~~~~~~~~~~~~~~~~~~~~~~~~~~~~~~~~~~~~~~~~~~~~~~~~~~~~~~~~~(c)~~~~~~~~~~~~~~~~~~~~~~(d)~~~~~~~~~~~~~~~~~~~~~~~~~~~~~~(e)
	\caption{ CTrGAN transfers the poses of the source to the target, while maintaining the natural gait of the target. From left  to right: (a) The source's image is converted  to (b) DensePose's~\cite{DBLP:conf/cvpr/GulerNK18} IUV format. (c) Our model translates the IUV of the source to the corresponding most natural IUV pose of the target by synthesizing a novel pose. (d) The generated pose is very similar (but not identical) to an exiting \emph{real} pose in the dataset. (e) The generated pose is rendered to a corresponding RGB image of the target.  }
	\label{fig:intro}
	\end{center}
\end{figure*}


Motion transfer methods aim to synthesize a video in which one individual is acting in accordance with that of a different individual in a given real video. A growing body of research has been conducted on this topic, which has led to the development of advanced detection~\cite{sabir2019recurrent,li2018ictu,chan2019dance} and enhanced motion transfer techniques~\cite{wang2018vid2vid,chan2019dance}. As a result of their mutually reinforcing relationship, motion transfer technology can produce convincingly realistic images and videos through deep learning-based manipulations. Using the whole-body motion transfer approach that directly works on unconstrained videos~\cite{chan2019dance} for gait transfer has the following key limitation: it attempts to replicate the precise movements of the source; rather, the goal of gait transfer is to translate the typical motions and appearances of the source into those of the target, adjusting for varied angles, paces, and shapes. To address this limitation we introduce CTrGAN for gait transfer. It transfers a series of poses from the source to the target while maintaining the natural movements of the target. Sources may vary in viewpoint, shape, and pace. It is based on Transformers \cite{NIPS2017_3f5ee243}, which have proven to be successful in translation tasks. Similar to NLP's models \cite{wolf-etal-2020-transformers,NEURIPS2020_1457c0d6}, each Transformer consists of an encoder and a decoder. As a result, we can successfully translate between the sequences of poses of the sources and the targets. Our Transformer model performs self- and cross- attention in time rather than in image space, capturing the dynamic of the object. In order to generate unseen natural poses of the target, our model is trained in an unsupervised manner on unpaired data. This is in contrast to prior whole-body motion transfer approaches that required paired data (e.g., \cite{wang2018fewshotvid2vid}). Figure~\ref{fig:intro} shows our method.

The quality of whole-body motion transfer is often evaluated in a supervised manner, based on the ability to approximate a pose and appearance of the target unseen during training but available in the test set. In our case, in addition to the appearance, the objective is to measure the translation of the gait pattern (i.e. the dynamic) from the source to a typical gait pattern of the target. In many cases, there are no one-to-one correspondences between each frame in the newly generated dynamic of the target and the existing ones already included in the dataset, and therefore it can not be evaluated directly in a supervised manner. We propose employing state-of-the-art gait recognition algorithms \cite{singh2018vision} to evaluate the quality of the gait transfer. The quality is determined by the accuracy with which newly generated poses are recognized as the target's gait. In order to provide accurate measurements in all scenarios, we use several different algorithms \cite{DBLP:conf/aaai/ChaoHZF19,DBLP:conf/cvpr/FanPC0HCHLH20,Lin_2021_ICCV}. As a way of assessing the appearances of unseen poses, Chamfer's distance~\cite{DBLP:conf/ijcai/BarrowTBW77} is used. 

Our model includes two networks: (a) CTrGAN, which translates the poses of the sources to the poses of the target, and (b) pose-to-appearance, which renders the appearance of each pose. For the latter network, we deploy an independent state-of-the-art existing approach.

This paper contributes by (a) introducing an approach for gait transfer from unconstrained videos in-the-wild, as well as evaluation metrics; (b) presenting Cycle Transformers GAN with temporal attention that can generate realistic gait patterns of the targets and a corresponding detector; and (c) demonstrating the effectiveness of our approach based on a standard gait recognition dataset, showing that it can generalize to unknown input sources, yielding the desired gait in an order of magnitude more cases than previous methods.


\section{Related work}
\label{Related_work}
\textbf{Pose-to-Pose/Appearance} A variety of methods have been introduced for the generation of video sequences of the target based on semantic input, including facial motion transfer \cite{DBLP:conf/cvpr/SiarohinLT0S19,kim2018deep,averbuch2017bringing,thies2016face2face} and whole-body motion transfer~\cite{chan2019dance,DBLP:conf/mipr/CormierMLMB21,wang2018fewshotvid2vid,wang2018vid2vid,ren2020human,wu2021human}. These methods are based on the ability to accurately estimate the pose \cite{DBLP:conf/cvpr/GulerNK18,cao2017realtime} and also on image-to-image translation models \cite{wang2018high,CycleGAN2017}. They are either explicitly trained for each source \cite{chan2019dance} or can be trained only once \cite{wang2018fewshotvid2vid} as our approach. In contrast to previous works, our goal is to generate the personalized gait pattern of the target to best match the gait of the source rather than to accurately imitate the original motion of the source.  Building on the recent advances, we employ \cite{chan2019dance,wang2018vid2vid} as our pose-to-appearance network where the input is the generated poses of the target and not the source's poses. Our experiments demonstrate the benefits of CTrGAN over the direct use of \cite{chan2019dance,wang2018vid2vid} for gait transfer. In the context of computer animation, \cite{aberman2020unpaired} introduced motion style transfer. However, they require a separation between the walking style and its content and 3D joint positions whereas we learn directly from the images containing the walking persons without disentangled representations.

\textbf{Gait Recognition}.  In recent years, various works have been proposed that use neural network models to identify people based on their gait \cite{DBLP:conf/aaai/ChaoHZF19,DBLP:conf/cvpr/FanPC0HCHLH20,DBLP:conf/icip/TeepeKGHHR21,DBLP:journals/sensors/CosmaR21}. GaitSet \cite{DBLP:conf/aaai/ChaoHZF19} considers the gait as a set consisting of independent frames and recognizes it based on a sequence of silhouette images. GaitGL \cite{Lin_2021_ICCV} relay on both global visual information and local region details and introduced attention between adjacent frames. GaitPart \cite{DBLP:conf/cvpr/FanPC0HCHLH20} uses a novel part-based model to characterize the gait. We use GaitSet, GaitGL, and GaitPart models to assess the quality of the generated video sequence. When the source in the video sequence is replaced with the target, the identified gait should be replaced as well. We show that for previous approaches, a gait is still readily associated with its source while using our approach it is considered to belong to the target.

\textbf{Visual Transformers}. Transformers~\cite{NIPS2017_3f5ee243} are proven architectures in the field of Natural Language Processing~\cite{NEURIPS2020_1457c0d6}, and several works have been done in recent years to adapt them to computer vision~\cite{46840,DBLP:conf/iclr/DosovitskiyB0WZ21,DBLP:conf/icml/HudsonZ21, Neimark_2021_ICCV}. The Transformer model is shown in \cite{NIPS2017_3f5ee243}  consists of two main components: an encoder and a decoder, which jointly process the input sequence, based on the self-attention mechanism. Early works  \cite{46840} adapted Transformers to the image domain. Even though this work demonstrated its ability only on very small images, it paved the way for broader works that addressed common challenges such as object detection \cite{DBLP:conf/eccv/CarionMSUKZ20} and classification \cite{DBLP:conf/iclr/DosovitskiyB0WZ21}. Recently, several works \cite{DBLP:conf/icml/HudsonZ21,DBLP:journals/corr/abs-2112-10762,jiang2021transgan} have been presented which show that Transformers can also be incorporated into the GAN architecture for image generation tasks. Unlike previous approaches, our method transfers motion between domains cyclically using unpaired data \cite{CycleGAN2017,DBLP:conf/eccv/BansalMRS18} and is based on attention in the temporal domain.

\section{Method}
\label{Method}
 \begin{figure*}[tb]
	\begin{center}
	\includegraphics[width=0.8\textwidth]{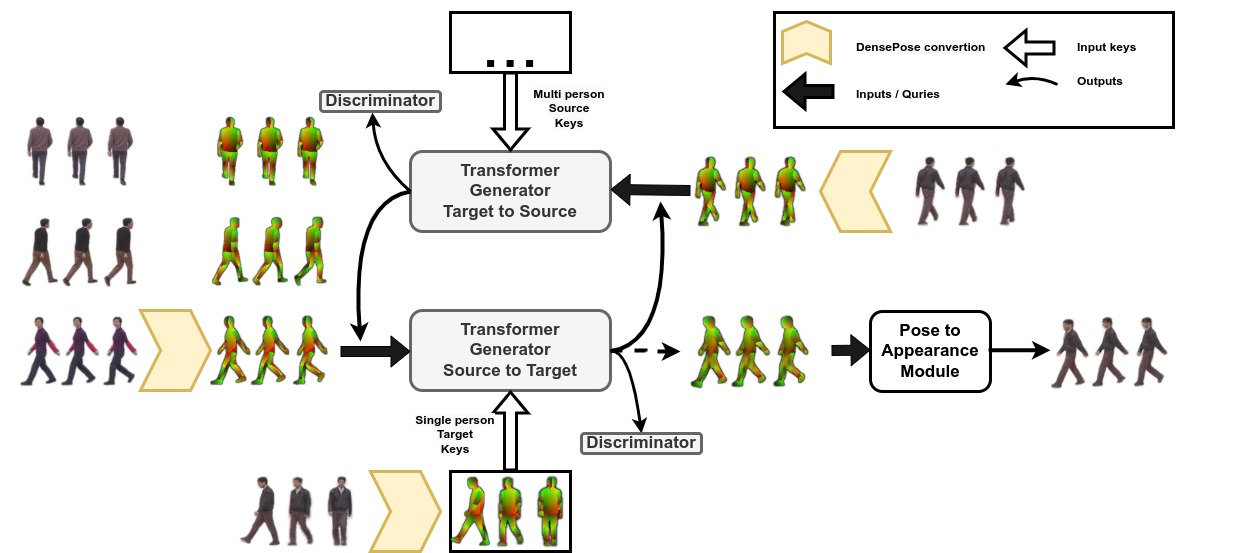}
	\caption{The generators of CTrGAN are based on transformers. The inputs to each generator are IUVA gait images from the training set and Keys. The outputs are natural gait poses.  See the text for further details.}
    \label{fig:CTrGAN-general}
	\end{center}
\end{figure*}

CTrGAN differs in the following ways from CycleGAN and standard Transformer-based architecture. First, unlike CycleGAN, it cycles between domains by using a series of images rather than individual images. Second, unlike CycleGAN and Transformers, the attention is on the temporal domain between consecutive images and not between patches of the same image. This allows us to incorporate the target's gait pattern into the source's gait pattern transfer process. Third, we do not use positional encoding due to the approximate cyclic pattern of gait. Figure~\ref{fig:CTrGAN-general} depicts a schematic illustration of our Natural Gait Retargeting approach.

\subsection{CTrGAN Architecture}

The Cycle Transormer GAN (CTrGAN) consists of three main ingredients: features extractors, Transformers, and a cyclic process. We denote $\mathcal{I} = \left \{\mathbf{I}_j \right \}_{j=1}^M$ as a collection of RGBA images and $\mathcal{P} = \left \{ \mathbf{P}_j \right \}_{j=1}^N$ as a collection of IUV images \cite{DBLP:conf/cvpr/GulerNK18}. $\mathbf{I}^{s_i}_j,\mathbf{P}^{s_i}_j$ and $\mathbf{I}^t_j,\mathbf{P}^t_j$ denote the corresponding $j^{th}$ image of the source and the target from the corresponding collection, respectively. In the following, we describe the details with respect to the target. Details regarding the source are derived in a similar manner. Below, the values in parenthesis represent those that we use in our implementation.

\begin{figure}
 	\centering
 	\includegraphics[width=0.9\columnwidth]{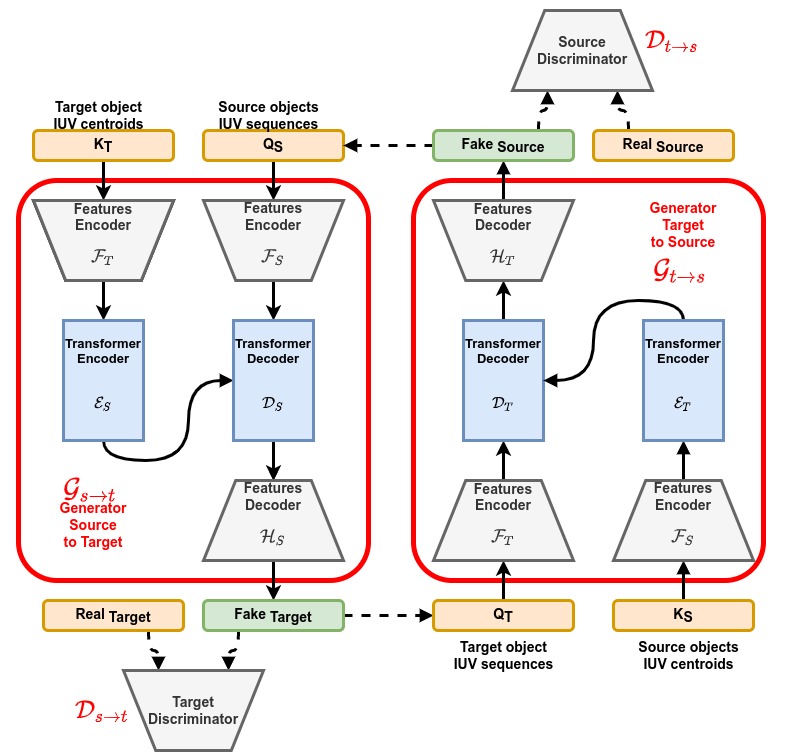}
 	
\caption{CTrGAN consists of two branches that are connected cyclically, feature encoders and decoders and Transformers which perform self- and cross-attention between the features.}

\label{fig:CTrGAN-arch}
\end{figure}   

\subsubsection{Transformers.} The Transformers follow the same architecture as presented in \cite{NIPS2017_3f5ee243}. Originally, Transformers were designed to handle sequences and consisted of two components, an encoder, and a decoder. The encoder is designed to handle information that remains constant throughout the series, while the decoder is designed to handle the continuous flow of information. The encoder and decoder consist of several chained attention blocks, and each receives three types of data as input: Keys, Values, and Queries (hereafter K, V, and Q).



\begin{figure}[tb]
\centering
\begin{subfigure}{1\columnwidth}
\centering
~~\\~~\\~~\\
	\includegraphics[width=0.7\columnwidth]{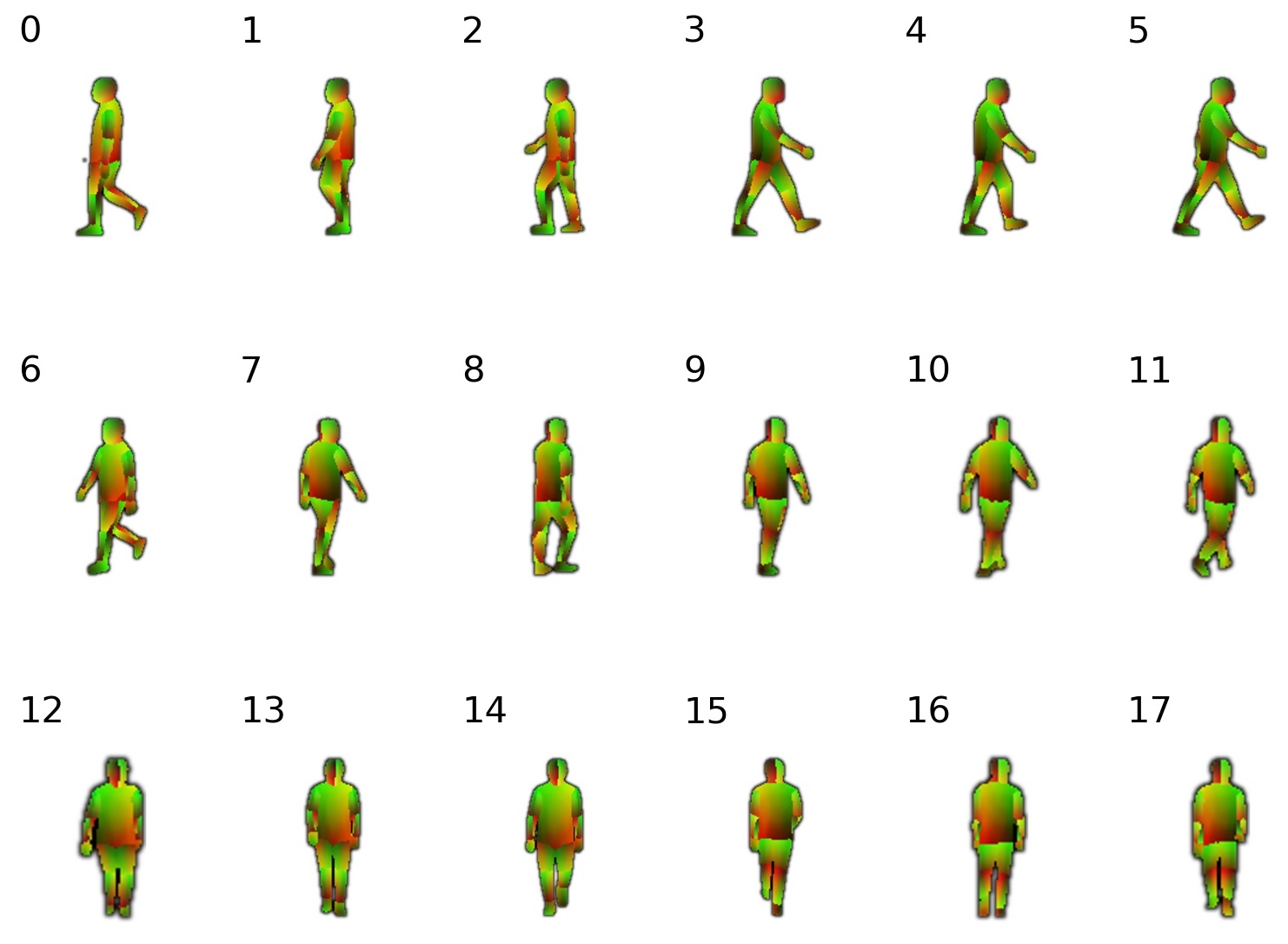}\\ ~~\\~~\\	
	\caption{Keys}
 	\label{fig:keys}
\end{subfigure}
\hfill
    \begin{subfigure}{1\columnwidth}
    \centering
    \includegraphics[width=0.9\columnwidth]{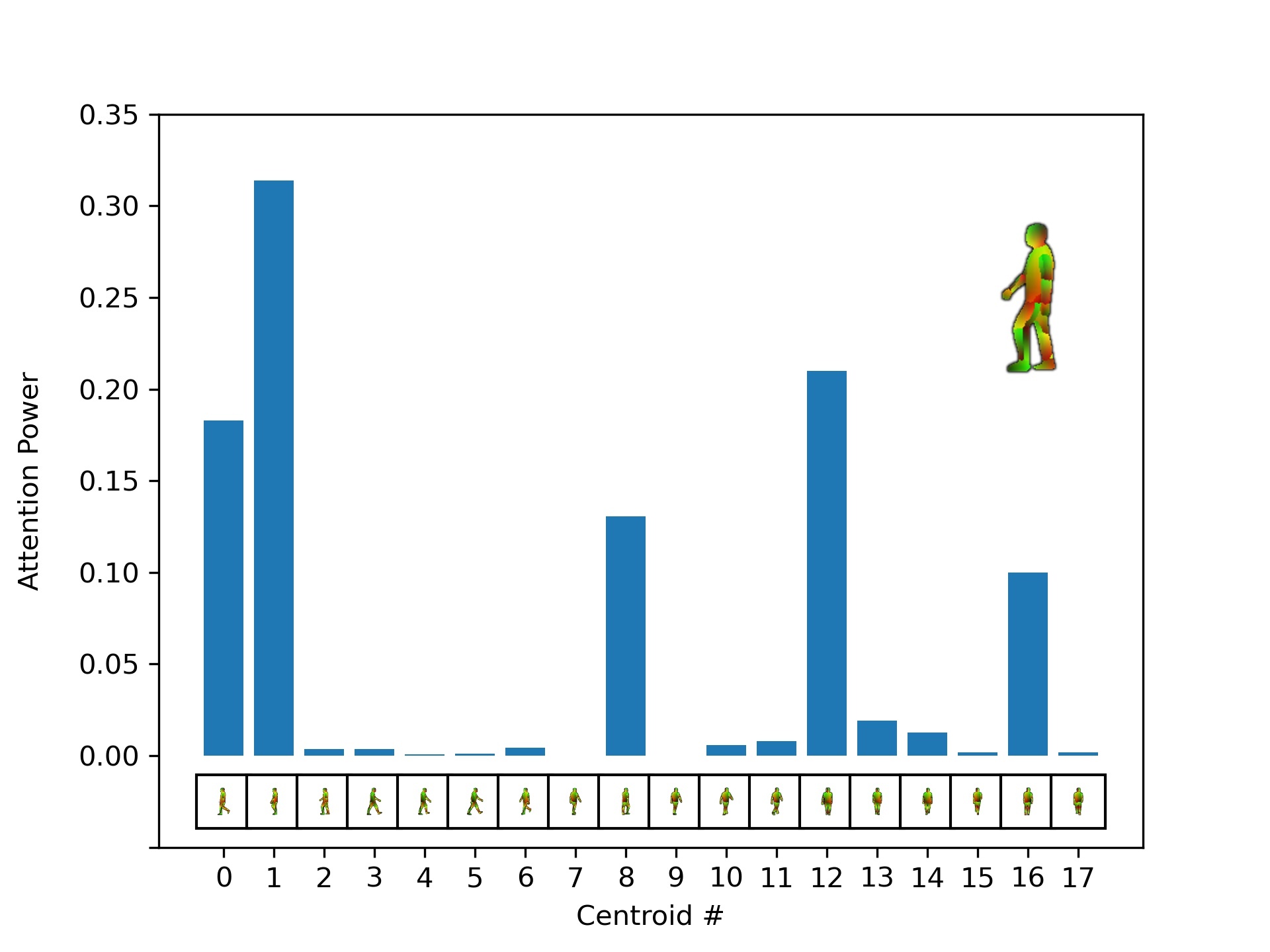} 	\caption{Attention}
    \label{fig:attention-visual}
    \end{subfigure}
\caption{(a) Samples of the Keys (centroids) $\mathbf{K}^t_k$ that were used. (b) A visual demonstration of our attention mechanism.} 
\label{fig:XXX}
\end{figure}

\begin{figure}[tb]
    \centering
    \begin{tabular}{cc}
        \includegraphics[width=0.9\columnwidth]{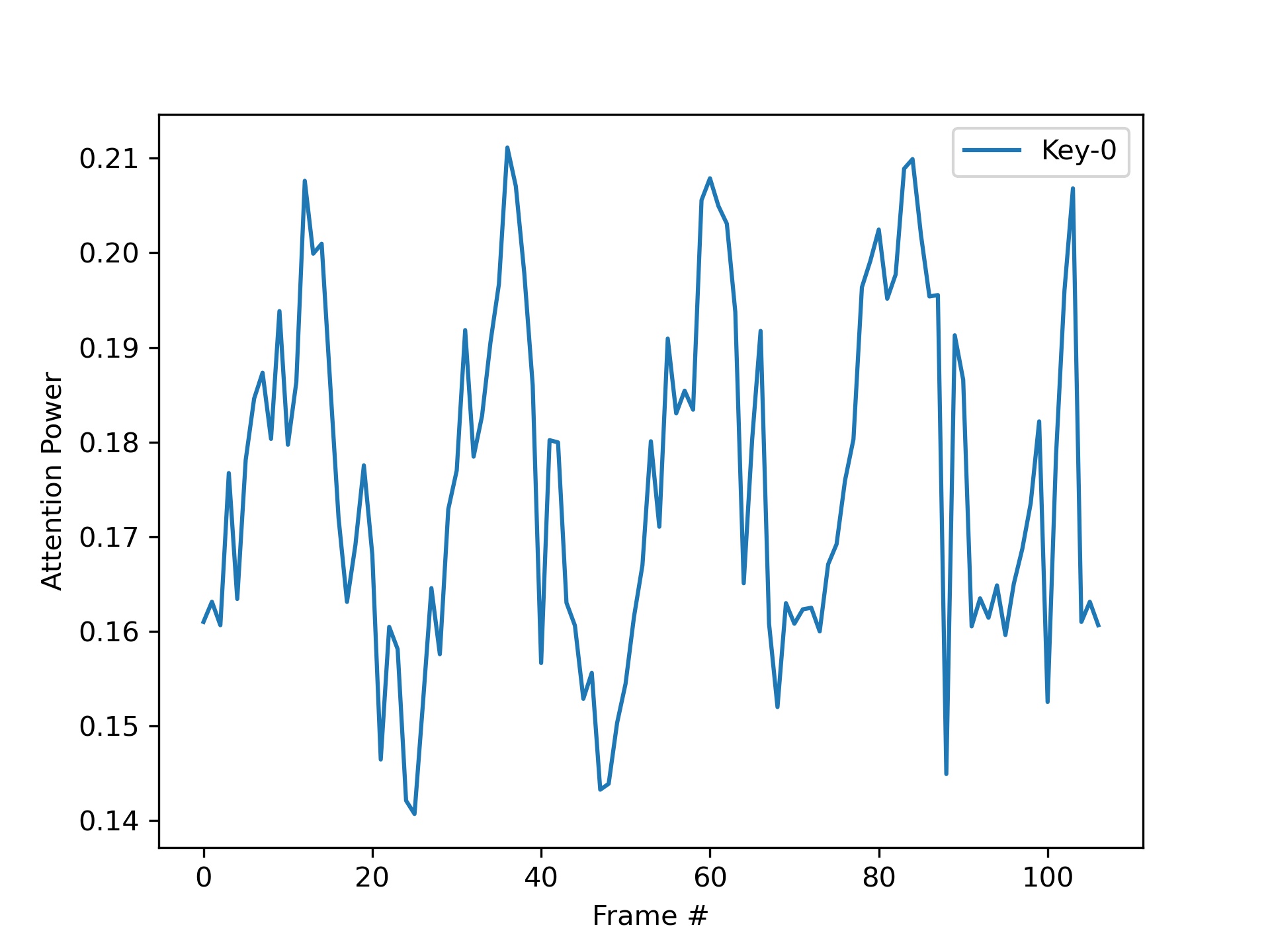}\\
        \includegraphics[width=0.9\columnwidth]{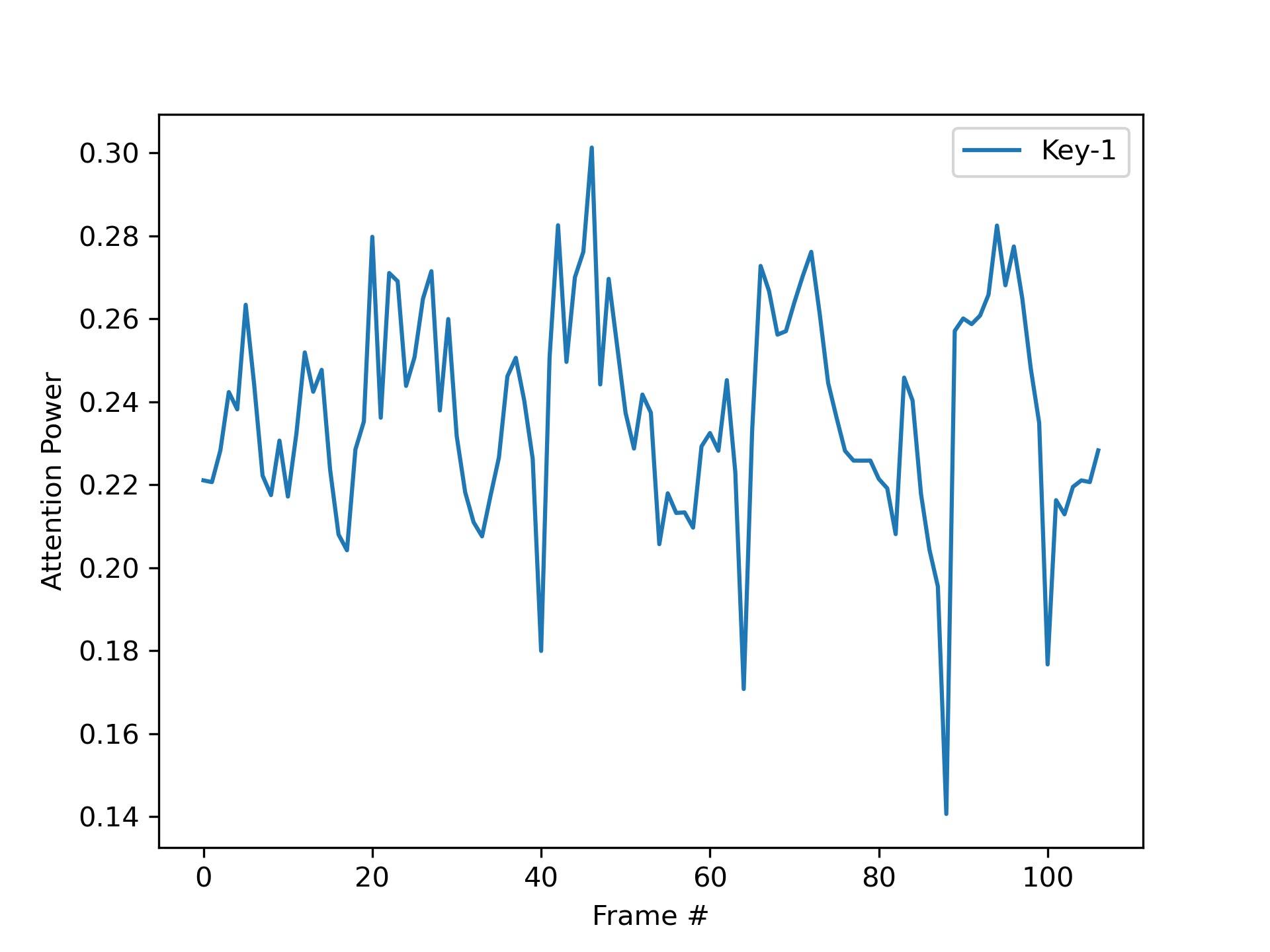}\\             
    \end{tabular}
    \caption{Attention power vs. time.}
    \label{fig:attn_time}
\end{figure}

\subsubsection{Keys} 
The Keys of the target are selected from the images' collection whose feature vectors are the closest to the cluster centers. Let $U_j$ denote the feature vector of image $j$:

\begin{equation}
	\{\mathbf{U}_j\} =  PCA(VGG_{16}(\{\mathbf{P}^t_j\}_{j=1..N}) , \textit{d}),
	\label{eq:features}
\end{equation}

where $VGG_{16}$ \cite{Simonyan15} are the features of the last layer (classifier) of the pre-trained VGG \footnote{Trained on ImageNet-1K} \cite{Simonyan15,NEURIPS2019_9015} and \textrm{d} is the PCA dimension ($\textrm{d}=100$).

$C_k$ denotes the center of the clusters as obtained by the K-means clustering:

\begin{equation}
	\{\mathbf{C}_k\} =  K_{means}( \{\mathbf{U}_j\} , \textrm{m}),
	\label{eq:centroids}
\end{equation}

where $\textrm{m}$ is the number of centroids ($\textrm{m}=18$). Finally, the Keys $\{\mathbf{K}^t_k\}$ are defined as:

\begin{equation}
    \mathbf{K}^t_k = \{\mathbf{P}^t_{p_k}\} \quad  , \quad p_k =\{ \underset{j}{\mathrm{argmin}} \| \mathbf{U}_j - \mathbf{C}_k \|_2 \}.
	\label{eq:QErr}
\end{equation}

Samples of the Keys that were used for one of the subjects can be seen in Figure~\ref{fig:keys}.

\subsubsection{Queries} Given an input sequence $\mathbf{P}^t$, the queries are a sliding window of $l_w$ consecutive frames ($l_w = 3$). We begin with a given frame and advance one sample at a time. During training, we choose a subsequence of length $L$ at random and begin with its first frame. We process the entire series from beginning to end during inference.


 

\subsubsection{Features Encoders and Decoders} Given an IUVA image $\mathbf{P}^{t}_j \in \mathbb{R}^{4\times H \times W}$, where IUVA is an IUV image with an additional alpha layer, we generates a feature tensor $\mathbf{U}^{t}_j \in \mathbb{R}^{256\times \frac{H}{64} \times \frac{W}{64}}$. The feature encoder $\mathcal{F}_T$ is a 5-layers CNN followed by 4 strides \textit{max-pooling} (see Supplementary for more details).

The CTrGAN model includes two pairs of encoders ($\mathcal{F}_T$ and $\mathcal{F}_S$), where each of the pairs shares weights. All four feature encoders (shown in Figure \ref{fig:CTrGAN-arch}) have the same structure. The $\mathcal{H}_T$ and $\mathcal{H}_S$ decoders are identical to the $\mathcal{F}_T$ encoder, except that they operate in the other direction. The discriminators  $D_{t \rightarrow s}$ and $\mathcal{D}_{t \rightarrow s}$  (Figure~\ref{fig:CTrGAN-arch}) are 5 -layers CNN (see Supplementary for more details).


\subsubsection{Cycle Transformer GAN}

Figure~\ref{fig:CTrGAN-arch} shows the architecture of CTrGAN . We denote the source images collections as $\Phi^{s_i} = \left \{ \mathbf{I}^{s_i}, \mathbf{P}^{s_i}  \right \}_{i=1..N }$ and the target image collection as $\Psi^t = \left \{ \mathbf{I}^t , \mathbf{P}^t \right \}$. We define two networks $\mathcal{G}_{s \rightarrow t}$ and $ \mathcal{G}_{t \rightarrow s}$. The first network adapts pose images from a variety of sources to pose images of the target, while the second network does the opposite. We denote $\widetilde{\mathbf{P}}^{t}$ and  $\widetilde{\mathbf{P}}^{s}$ as the outputs of the networks:

\begin{equation}
 \left \{ \widetilde{\mathbf{P}}^{t}  \right \}= \mathcal{G}_{s \rightarrow t}\left (  \{  \mathbf{K}^{t} \} ,\left \{  \mathbf{P}^{s_i} \right \}  \right ),
 \end{equation}
 \begin{equation}
 \left \{ \widetilde{\mathbf{P}}^{s_i}  \right \} = \mathcal{G}_{t \rightarrow s}\left (  \{  \mathbf{K}^{s_i} \} ,\left \{  \mathbf{P}^t \right \}  \right ).
  \end{equation}
  
For brevity, images' indices have been omitted. Using the pose images of the source's gait pattern, our method attempts to generate pose images of the target while preserving its gait pattern. This domain adaptation is accomplished by composing both $\mathcal{G}_{s \rightarrow t}$ and $ \mathcal{G}_{t \rightarrow s}$ in a cyclic manner as shown in the following:

  \begin{equation}
 \widetilde{\mathbf{P}}^{s_i}   = \mathcal{G}_{t \rightarrow s}\left ( \mathbf{K}^{s_i}, \mathcal{G}_{s \rightarrow t}\left (   \mathbf{K}^{t} ,  \mathbf{P}^{s_i}  \right ) \right ),
 \label{eq:G-S2T}
 \end{equation}
 \begin{equation}
  \widetilde{\mathbf{P}}^{t} = \mathcal{G}_{s \rightarrow t}\left ( \mathbf{K}^{t}, \mathcal{G}_{t \rightarrow s}\left (   \mathbf{K}^{s_i} ,  \mathbf{P}^t  \right ) \right ).
  \label{eq:G-T2S}
 \end{equation}

The output pose image $\widetilde{\mathbf{P}}^{t} $ (\textit{i.e.} $\mathcal{G}_{s \rightarrow t}\left (   \mathbf{K}^{t} ,  \mathbf{P}^{s_i}  \right )$)are used to generate the requested appearance by pose-to-appearance network $\mathcal{G}_{M}$: 
\begin{equation}
\widetilde{\mathbf{I}}^t_i = \mathcal{G}_{M}\left ( \mathbf{P}^t_i \right ).
  \label{eq:G-M}
\end{equation}

\subsubsection{Self and Cross-Attention}

The attention layer at the attention block is one of the core components of the Transformer. A detailed explanation of the self- and cross-attention and visualizations can be found in the Supplementary Materials. Figure~\ref{fig:attention-visual} visualizes the attention mechanism.

\subsubsection{Gait Cycle}

We do not use positional encoding due to the cyclic pattern of gait~\cite{silva2020basics}. The gait cycle can be defined as the time interval between two successive occurrences of one of the repetitive phases of locomotion \cite{alamdari2017review}. Here, we demonstrate the periodicity of the movement as expressed by the cross-attention patterns along time for the Keys. Figure~\ref{fig:attn_time} shows the cross-attention of Key 0 (top) and Key 1 (bottom) over time. As can be seen, the gait cycle is evident.

\subsection{Optimization and Loss Functions}
We use the following loss functions in our training:
\begin{equation}
	\ell_{cycle} = \lambda_{idt} \ell_{idt} + \lambda_{adv} \ell_{adv} + \lambda_{cyc} \ell_{cyc} +  \lambda_{per} \ell_{per} ,
\end{equation}

where $\lambda_{idt}$, $\lambda_{adv}$ , $\lambda_{cyc}$ and $ \lambda_{per}$ are the weights of the losses.  The identity loss function $\ell_{idt}$ is used to ensure that the cyclic mapping preserves the mapping from a pose image to itself.  We use $\mathcal{L}_1$ loss function. 
\begin{equation}
	\ell_{idt} = \mathcal{L}_1\left ( \mathcal{G}_{s \rightarrow t}\left ( \mathbf{P}^t \right ) ,  \mathbf{P}^t\right ) +  \mathcal{L}_1\left ( \mathcal{G}_{t \rightarrow s}\left ( \mathbf{P}^s \right ) ,  \mathbf{P}^s\right ).
\end{equation}

Following CycleGAN \cite{CycleGAN2017} and ReCycleGAN \cite{DBLP:conf/eccv/BansalMRS18} we use the same adversarial loss $\ell_{adv}$, in which two discriminator networks $\mathcal{D}_{s \rightarrow t}$ and $ \mathcal{D}_{s \rightarrow t}$ are learned as a part of the training process.

In the same manner as GAN \cite{NIPS2014_5ca3e9b1} architectures, we use the generator and discriminator $\mathcal{G}_{s \rightarrow t}$, $\mathcal{G}_{t \rightarrow s}$ $\mathcal{D}_{s \rightarrow t}$  and $\mathcal{D}_{t \rightarrow s}$. 
In our training process, we use the $\mathcal{L}_2$ as the objective loss function of the adversarial loss $\ell_{adv}$.
\begin{equation}
	\begin{split}
		\ell_{adv} = \mathcal{L}_2\left ( \mathcal{D}_{s \rightarrow t} \left (\mathcal{G}_{s \rightarrow t}\left ( \mathbf{P}^s \right )  \right ) ,  \mathbb{0} \right ) 
		+ \mathcal{L}_2\left ( \mathcal{D}_{s \rightarrow t} \left ( \mathbf{P}^t  \right ) ,  \mathbb{1}\right )\\
		  +  \mathcal{L}_2\left (   \mathcal{D}_{t \rightarrow s} \left (\mathcal{G}_{t \rightarrow s}\left ( \mathbf{P}^t \right ) \right ) ,  \mathbb{0}
		\right ) +  \mathcal{L}_2\left (   \mathcal{D}_{t \rightarrow s} \left ( \mathbf{P}^s \right ) ,  \mathbb{1}
		\right )
	\end{split},
\end{equation}

where $\mathbb{0}$ and $\mathbb{1}$ are matrices of zeros and ones with the same dimensions as $\mathcal{D}_{s \rightarrow t} \left (\mathcal{G}_{s \rightarrow t}\left ( \mathbf{P}^s \right ) \right )$ and $ \mathcal{D}_{t \rightarrow s} \left (\mathcal{G}_{t \rightarrow s}\left ( \mathbf{P}^t \right ) \right )$.

The cycle loss function $\ell_{cyc}$ which is the main core of the cycle GAN process is defined as:
\begin{equation}
	\begin{split}
	\ell_{cyc} = \mathcal{L}_2\left ( \mathcal{G}_{t \rightarrow s} \left (\mathcal{G}_{s \rightarrow t}\left ( \mathbf{P}^s \right ) \right ) ,  \mathbf{P}^s\right ) \\ +  \mathcal{L}_2\left (   \mathcal{G}_{s \rightarrow t} \left (\mathcal{G}_{t \rightarrow s}\left ( \mathbf{P}^t \right ) \right ) ,  \mathbf{P}^t\right ).
	\end{split}
\end{equation}

The perceptual loss $\ell_{per}$ measures the difference between feature vectors of a real image and a generated one:

\begin{equation}
	\begin{split}
	\ell_{per} = \mathcal{L}_1\left ( VGG_{16}  \left (\mathcal{G}_{t \rightarrow s} \left (\mathcal{G}_{s \rightarrow t}\left ( \mathbf{P}^s \right ) \right )  \right ),  VGG_{16}  \left (\mathbf{P}^s\right )  \right ) \\ +  \mathcal{L}_1\left (  VGG_{16}  \left ( \mathcal{G}_{s \rightarrow t} \left (\mathcal{G}_{t \rightarrow s}\left ( \mathbf{P}^t \right ) \right )\right ) ,  VGG_{16}  \left (\mathbf{P}^t\right ) \right )
	\end{split}.
\end{equation}

We use a pre-trained VGG model \cite{Simonyan15,NEURIPS2019_9015} to extract the feature vectors. VGG model is pre-trained on three layers of RGB, whereas we use four layers (IUVA), we therefore measured the perceptual loss of the IUV and the alpha channels separately.

\section{Experiments}
\label{results}
\subsection{Dataset}


We use CASIA-A \cite{DBLP:journals/pami/WangTNH03}, which is a widely-used gait recognition dataset \cite{DBLP:conf/cores/KhanLFG17,Ekinci2006HumanIU,Arantes2010HumanGR,10.1145/3490235}. It includes twenty subjects. Each subject has twelve image sequences that were captured from three different viewpoints, resulting in four instances for each of the viewpoints (denoted by 001, 002, 003 and 004). Overall, there are a total of 240 video sequences with a resolution of 352x240 and 19,139 images.




 
 \begin{figure}
 	\centering
 	\begin{tabular}{cc}
 		\includegraphics[trim=0 0 8 0, clip,width=0.4\linewidth]{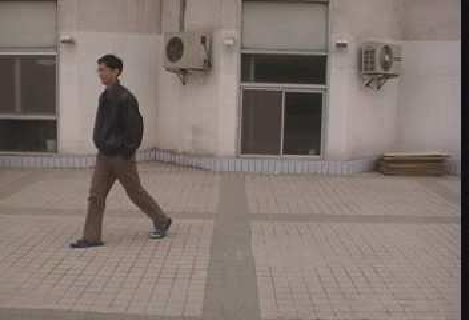} &
 		\includegraphics[width=0.27\linewidth]{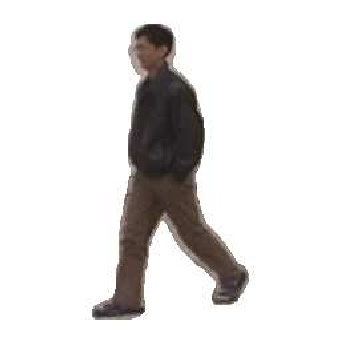} \\
 		\includegraphics[width=0.27\linewidth]{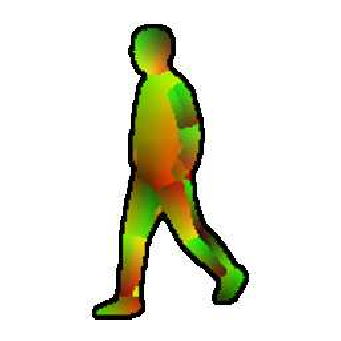} &
 		\includegraphics[width=0.27\linewidth]{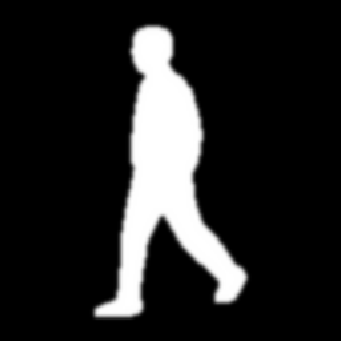}
 	\end{tabular}

 	\caption{Pre-processing procedure - From left to right: 1. Original image. 2. Cropped and centered image. 3. Pose image (IUV - DensePose format). 4. Masked image, created from the I part of the IUV image}
 	\label{fig:casia-samples} 
 \end{figure}
 

We remove the background from the original CASIA-A images by using DensePose  \cite{DBLP:conf/cvpr/GulerNK18}. The input to our model is four-channel images (RGBA). In each frame, we extract the binary mask for the subject and attach it to the RGB image as an alpha channel. The images are cropped and centered around the object to create 256x256 canvases. See Figure~\ref{fig:casia-samples} for examples.



\subsection{Implementation Details}



The networks are implemented using Pytorch \cite{NEURIPS2019_9015} and were trained on a single NVidia 2080Ti GPU. We train the model with Adam optimizer with $\beta1=0.5$ $\beta2=0.999$ over 20 epochs. The initial learning rate is set to  $2e-4$, for 5 epochs followed by a linear decay to zero over 15 more epochs. The same configurations and parameters are used for all models ($\mathcal{F_T}$,$\mathcal{F_S}$,$\mathcal{H_T}$,$\mathcal{H_S}$,$\mathcal{E_S}$,$\mathcal{D_S}$ ,$\mathcal{E_T}$,$\mathcal{D_T}$ and $\mathcal{D}_{s \rightarrow t}$,$\mathcal{D}_{t \rightarrow s}$). In order to represent temporal relations more effectively, we use three consecutive frames as a mini-batch. Our augmentations include a small magnification (from 256 to 272) and random cropping.

A detailed description of our architecture can be found in the Supplementary Materials.


\subsection{Baselines}

The baselines are state-of-the-art methods for motion retargeting, V2V \cite{wang2018vid2vid} and EDN \cite{chan2019dance}. The models are adapted to include an alpha channel as well. EDN is adjusted to work with IUVA (IUV +  alpha channel), whereas the V2V model is already optimized for DensePose images, so only one more channel is needed. We train V2V and EDN according to their protocol with the default parameters. 

We evaluate the following approaches to assess CTrGAN's contribution: (a) direct - using trained baselines to map directly from pose to appearance. (b) ours - we use CTrGAN to generate the pose images, then use the baselines to render the appearance.

\subsection{Metrics}
\textbf{Gait quality.} We evaluate our results by gait recognition models GaitSet \cite{DBLP:conf/aaai/ChaoHZF19}, GaitGL  \cite{Lin_2021_ICCV}  and  GaitPart \cite{DBLP:conf/cvpr/FanPC0HCHLH20}, implemented by the OpenGait \cite{OpenGait2022} package. 


We report  \textbf{target-accuracy} - the percentages of times the gait recognition model identifies the generated gait as the target's gait. Given a set of reference videos $\{\mathbf{I^{s_i}}\}$ and a generated video $\mathbf{I^g}$, the goal is to find the reference video in which the gait pattern is the most similar. For that, references videos are ranked according to their distance from $\mathbf{I^g}$:

\begin{equation} 
	D_{s_i,g}=\| \mathcal{M}(\mathbf{I^{s_i}}) -  \mathcal{M}(\mathbf{I^g})\|_2,
	\label{eq:gait-distance}
\end{equation}

where $\mathcal{M}$ is the specific model. The most similar reference video is considered as the one with the highest top-3 (minimum distance) frequency $\mathbf{I^{s_i*}}$. The identified gait is of the subject in $\mathbf{I^{s_i*}}$ and the recovered distance is $D_{s_i*,g}$. In our case the generated video $\mathbf{I^g}$ is $\mathcal{G}_{M} \left( \mathcal{G}_{s \rightarrow t}\left( \mathbf{P^g} \right) \right)$, where $\mathcal{G}_M$ is the pose to appearance model that is in use (\textit{e.g.} V2V).



We train the models on all the subjects and half of the CASIA-A videos.


\textbf{Appearance quality.}  We use the following metrics to evaluate our appearance quality: Inception Score (IS)\cite{NIPS2016_8a3363ab},  Structural Similarity (SSIM) \cite{DBLP:journals/tip/WangBSS04}, Perceptual Image Patch Similarity (LPIPS), \cite{DBLP:conf/cvpr/ZhangIESW18} and Frechet Inception Distance (FID) \cite{DBLP:conf/nips/HeuselRUNH17}.

FID and IS metrics measure statistical differences between sets of images and not directly between individual images. However, SSIM and LPIPS evaluate the generated image based on a single, ground truth image. Due to the fact that the generated synthetic image $\mathbf{\widetilde{I}^t_i}$ in our test set can be derived from an unseen source's pose, a ground-truth image $\mathbf{{I}^t_i}$ is not always available. We therefore use the Chamfer Distance \cite{DBLP:conf/ijcai/BarrowTBW77} to recover the nearest ground truth image from the reference video sequence $\{I^r_k\}$:
\begin{equation}
	E_{CD}=\dfrac{1}{N}\sum_{i=1}^{N} \min_{k} Q(\mathbf{\widetilde{I}^t_i},\mathbf{I^r_k}),
	\label{eq:Chamfer}
\end{equation}

where $Q$ represents our quality metrics, SSIM or LPIPS. All methods are compared using $E_{CD}$, both ours and others.

\textbf{Pose retargeting quality.}  To estimate the extent to which the basic motion of the source is transferred to the target, the intersection over union (IoU) of the binary silhouettes of the source and target is used. The binary silhouette $\mathcal{S}(\mathbf{P^{A}_k})$ of subject $A$ in frame $k$ is generated from the alpha channel of its IUVA image $I_{\alpha}$ with a threshold of  $\frac{max(\mathbf{I_{\alpha}}
) - min(\mathbf{I_{\alpha}})}{2}$.

Given two sequences $A$ and $B$ with $M$ and $N$ frames respectively, we calculate the average IoU between them according to:

\begin{equation}
R(A,B)=\frac{1}{MN} \sum_j^M \sum_k^N IoU(\mathcal{S}(\mathbf{P^{A}_j}), \mathcal{S}(\mathbf{P^{B}_k})).
\label{eq:Rs}
\end{equation}

We report R where A is the generated target and B is the source, and when A is the generated target and B is its original sequence.

\subsection{Experiments}

\subsubsection{CTrGAN successfully generates the gait of the target.}

We train our model on thirteen subjects and use the remaining seven for testing. For each subject, both the training and test sets include two video sequences. In this way, we can include in our evaluation cases where sources were not available during training.


For the test set, we generate video sequences for the trained subjects. We deploy all subjects in the dataset as sources, including those not included in the training set. We test the ability to identify the generated gait as the target's gait by the gait recognition models, before and after applying our approach. Our results demonstrate that our approach can generate a more realistic gait for the target by an order of magnitude than previous methods.

Table~\ref{tab:gait-metric-target} presents our main results, the target-accuracy for V2V and EDN. In all the tables, bold represents the best result. The top of the table shows the baselines applied directly to the pose of the source. At the bottom of the table, CTrGAN generates the poses of the target before applying the baseline methods. It can be seen that for all the methods, CTrGAN significantly improves the ability to generate the natural gait of the target. All models have failed to recognize the gait of the target in the case of V2V without CTrGAN. This implies that indeed, V2V can accurately mimic the movements of the source by the target, in accordance with its original goal. On average, the generated gait rendered by V2V is approximately $17$ times more likely to match the target's gait when using our method. 

\begin{table}


\centering
		\begin{tabular}{|c|c|c|c|c|}
		\hline
		    Method &  Model &  GaitPart  &     GaitSet &  GaitGL  \\
			\hline
              \multirow{ 2}{*}{-}
            &   EDN &  16.94  & 29.44    &   16.67     \\             
            &   V2V &      3.89&   3.61   &    4.17   \\ 
             \hline	\hline	

             \multirow{ 2}{*}{ours} 
              & EDN &   18.89  & \textbf{62.78}     &   36.39   \\             
              & V2V &     \textbf{84.72}  & 56.67   &   \textbf{68.06 }  \\

			\hline	
		\end{tabular}
\caption{The target-accuracy $\uparrow$. The top row is before and the bottom row is after applying CTrGAN to generate the poses. Our approach significantly improves the ability to generate the target's gait, up to approximately $\times$21 than existing methods (GaitPart+V2V).}
\label{tab:gait-metric-target}
\end{table}


\begin{figure}
\centering
 	\begin{tabular}{cc}
    \begin{subfigure}{0.4\columnwidth}
    	\includegraphics{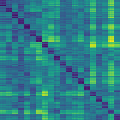}
    	\caption{Before}
     	\label{fig:gait-metric-direct}
    \end{subfigure} &
 		\    \begin{subfigure}{0.4\columnwidth}
    	\includegraphics{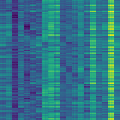}
 	    \caption{After}
    \label{fig:gait-metric-ctrgan}
    \end{subfigure}\\

 	\end{tabular}
	
\caption{GaitSet's distance matrix for subject three (the target) in the training set before and after applying our method. The darker the color the lower the value. It can be seen that before deploying our model, GaitSet easily distinguishes between the generated and real gait and can identify the true sources. After applying our approach, GaitSet identifies for most of the cases the generated gait as the real gait of subject three. }  
\label{fig:gait-metric}
\end{figure}

 \begin{table*}
\centering
\begin{tabularx}{0.8\linewidth}{|l|L|L|L|c|} 

    \hline
    \multicolumn{1}{|c|}{} &\multicolumn{3}{c|}{Features } &\multicolumn{1}{c|}{ } \\
     \cline{2-4}
     
		 Model & Attention & Encoder & Decoder & Target-accuracy $\uparrow$ \\
		  & mechanism & self-attention & self-attention & \\
		\hline
 Cycle Only &\xmark&\xmark&\xmark &     5.28   \\ 
+ Attention  &\checkmark&\checkmark&\xmark&   66.21   \\ 
+ Time-Attention &\checkmark&\checkmark&\checkmark&  69.82 \\ 
 		\hline
		\end{tabularx}

\caption{The target-accuracy  of several CTrGAN configurations.}

\label{tab:gait-metric-config}
\end{table*}

Figure \ref{fig:gait-metric} shows the GaitSet's distances for subject number three (the target) with respect to all the twenty subjects in the dataset (the sources). Dark colors represent low values, whereas light colors represent high values. The lower the distance, the more similar the gait in the reference sequence is to that in the generated sequence. Figure \ref{fig:gait-metric-direct} presents the distance matrix for the V2V method in the direct approach. It can be seen that GaitSet is able to accurately recognize the sources for all the generated sequences. Figure~\ref{fig:gait-metric-ctrgan} presents the results for V2V after applying our CTrGAN to generate the poses. For the vast majority of the sequences, GaitSet recognizes the generated gait of the target (subject three) as the real gait.


\begin{table}[tb]

	\begin{center}

		\begin{tabular}{|c|c|c|c|c|c|}
			\hline
			\multirow{ 2}{*}{Method} & \multirow{ 2}{*}{Model}  &  SSIM   & LPIPS   &  FID$\downarrow$  & IS$\downarrow$ \\
			&  & [CD] $\uparrow$& [CD]$\downarrow$  & &  \\
			\hline\hline
               \multirow{ 3}{*}{-} 
               & EDN &         0.890 &     0.072 & 55.79 & 0.0025 \\
                &V2V  &    0.901 &     0.063 &   53.131 & 0.0010 \\                
                \hline\hline

                \multirow{3}{*}{ours} 
                &   EDN &           0.870 &     0.101 &    83.67 & 0.0030 \\
                & V2V &    \textbf{0.909} &     \textbf{0.055} &    \textbf{52.89} & \textbf{0.0009}\\
             \hline
		\end{tabular}
				\caption{Appearance quality.}
	    \label{tab:NN-Q}

	\end{center}
\end{table}

Table \ref{tab:NN-Q} shows the appearance quality of the different approaches when deploying pose to appearance networks, with and without CTrGAN. Without CTrGAN, the appearance metrics are similar across the different methods. CTrGAN slightly increases the appearance quality for V2V but overall the metrics are comparable. It is expected as the key contribution of CTrGAN is the generation of poses that can naturally be attributed to the target rather than improving the rendering mechanism of an existing pose. A possible explanation for the slight improvement could be that the generated poses by CTrGAN match more naturally the target's appearances that need to be rendered.

The average pose retargeting quality obtained by applying \cref{eq:Rs} to all the sequences of the generated targets and the sequences of their corresponding sources is 0.8677. The average pose retargeting quality obtained by applying \cref{eq:Rs} to all the sequences of the generated targets and their original sequences before applying CTrGAN is 0.6325. It is evident that CTrGAN transforms the pose's silhouette of the target to be very similar to the source poses' silhouette.

In the Supplementary Material, we show that existing motion transfer methods retain the gait pattern of the source.


\subsubsection{Temporal attention improves accuracy.}

Table \ref{tab:gait-metric-config} shows the effect of the different components of CTrGAN on the final results. We employ V2V as our pose-to-appearance network and evaluate the average target-accuracy over all gait recognition models. It can be seen that CycleGAN architecture on its own is not sufficient in order to generate natural poses of the target. Adding encoder self-attention and cross-attention between the image sequence and the keys using the decoder produces significantly more natural poses. A further improvement is obtained when the decoder self-attention is added, which takes advantage of the temporal relations within the sequence. Additional detailed comparisons can be found in the Supplementary Materials.

\section{Detecting Gait Transfer}


It is imperative to carefully consider the implications of our method, particularly in light of recent events that are occurring, where misinformation is being used systematically. In introducing the gait transfer problem, we hope to increase awareness of this important issue. We investigate methods for detecting generated gait transfer videos as a first step towards preventing misuse of our approach. In order to identify videos created by our model, we train an appearance-based detector. Choosing an appearance-based detector is practical since the key contribution of our approach is the generation of natural gait patterns whereas the appearance quality is comparable to that of existing motion transfer methods. Our detector classifies videos as either real or generated. The dataset includes both original and generated images of a walking person in an outdoor environment. We use 75 percent of the subjects for training and 25 percent for testing. The deployed model is ResNet152\cite{he2016deep}. Using transfer learning, the last FC layer of the model trained on ImageNet is replaced with one that is adapted for two classes, which is fine-tuned on the training data. We achieve an average detection accuracy of 96.2\% for held-out target subjects.

\section{Conclusion}
\label{conclusion}

We introduce a novel approach for gait transfer based on unconstrained videos in-the-wild. We propose quantifiable metrics to better evaluate the quality of the transfer. We present CTrGAN, a novel Transformer-based architecture. Our model cycles between domains by using a series of images and includes self-, cross-, and temporal-attention. We introduce an appearance-based detector and show that it can be highly accurate. Using our approach, we obtained state-of-the-art results.





%
%
{\small
\bibliographystyle{ieee_fullname}
\bibliography{gait_transform}
}

\end{document}